\documentclass{article}

\usepackage{microtype}
\usepackage{graphicx}
\usepackage{subfigure}
\usepackage{booktabs} 

\usepackage{hyperref}

\usepackage[accepted]{icml2025}

\usepackage{amsmath}
\usepackage{amssymb}
\usepackage{mathtools}
\usepackage{amsthm}

\usepackage[capitalize,noabbrev]{cleveref}

\theoremstyle{plain}
\newtheorem{theorem}{Theorem}[section]
\newtheorem{proposition}[theorem]{Proposition}
\newtheorem{lemma}[theorem]{Lemma}
\newtheorem{corollary}[theorem]{Corollary}
\theoremstyle{definition}
\newtheorem{definition}[theorem]{Definition}
\newtheorem{assumption}[theorem]{Assumption}
\theoremstyle{remark}
\newtheorem{remark}[theorem]{Remark}

\usepackage[textsize=tiny]{todonotes}

\icmltitlerunning{DAWM: Diffusion Action World Models for Offline Reinforcement Learning via Action-Inferred Transitions}

\begin{document}

\twocolumn[
\icmltitle{DAWM: Diffusion Action World Models for Offline Reinforcement Learning via Action-Inferred Transitions}



\icmlsetsymbol{equal}{*}

\begin{icmlauthorlist}
\icmlauthor{Zongyue Li}{equal,yyy,comp}
\icmlauthor{Xiao Han}{equal,yyy}
\icmlauthor{Yusong Li}{yyy}
\icmlauthor{Niklas Strau\ss}{yyy,comp}
\icmlauthor{Matthias Schubert}{yyy,comp}

\end{icmlauthorlist}

\icmlaffiliation{yyy}{Department of Computer Science, University of Munich, Munich, Germany}
\icmlaffiliation{comp}{Munich Center for Machine Learning (MCML), Munich, Germany}

\icmlcorrespondingauthor{Matthias Schubert}{schubert@dbs.ifi.lmu.de}
\icmlcorrespondingauthor{Zongyue Li}{li@dbs.ifi.lmu.de}

\icmlkeywords{Machine Learning, ICML}

\vskip 0.3in
]



\printAffiliationsAndNotice{\icmlEqualContribution} 
\begin{abstract}
Diffusion-based world models have demonstrated strong capabilities in synthesizing realistic long-horizon trajectories for offline reinforcement learning (RL). However, many existing methods do not directly generate actions alongside states and rewards, limiting their compatibility with standard value-based offline RL algorithms that rely on one-step temporal difference (TD) learning. While prior work has explored joint modeling of states, rewards, and actions to address this issue, such formulations often lead to increased training complexity and reduced performance in practice. We propose \textbf{DAWM}, a diffusion-based world model that generates future state-reward trajectories conditioned on the current state, action, and return-to-go, paired with an inverse dynamics model (IDM) for efficient action inference. This modular design produces complete synthetic transitions suitable for one-step TD-based offline RL, enabling effective and computationally efficient training. Empirically, we show that conservative offline RL algorithms such as TD3BC and IQL benefit significantly from training on these augmented trajectories, consistently outperforming prior diffusion-based baselines across multiple tasks in the D4RL benchmark.
\end{abstract}

\section{Introduction}

Recent advances in world models have spurred increasing interest in using learned environment models to facilitate planning in offline reinforcement learning (RL). To this end, various sequence modeling approaches have emerged as effective frameworks for training such world models, including recurrent architectures \cite{hafner2020dreamcontrollearningbehaviors, hafner2022masteringataridiscreteworld, hafner2024masteringdiversedomainsworld} that model temporal dynamics and reward models, transformer-based models that learn structured representations of state, action, and reward sequences to model long-term dependencies \cite{chen2021decisiontransformerreinforcementlearning, janner2021offlinereinforcementlearningbig, micheli2023transformerssampleefficientworldmodels, robine2023transformerbasedworldmodelshappy}, and diffusion-based generative models \cite{ajay2023conditionalgenerativemodelingneed,janner2022planningdiffusionflexiblebehavior,ding2024diffusionworldmodelfuture,zhou2024diffusionmodelpredictivecontrol, li2025beyond} that can generate entire trajectories in a single forward pass rather than autoregressively predicting each timestep, thus mitigating compounding error and enabling more coherent long-horizon planning \cite{ding2024diffusionworldmodelfuture}.

Extending prior work, we explore the use of diffusion-based world models to generate long-horizon synthetic trajectories suitable for offline reinforcement learning. An expanding body of works \cite{alonso2024diffusionworldmodelingvisual,ding2024diffusionworldmodelfuture, janner2022planningdiffusionflexiblebehavior, jackson2024policyguideddiffusion, rigter2024worldmodelspolicyguidedtrajectory} has demonstrated strong capabilities in modeling complex, multimodal dynamics across diverse offline datasets, as summarized in Table \ref{difftable}. 


\begin{table*}[t]
\caption{A comparison of diffusion-based world models for RL.}
\label{difftable}
\vskip 0.15in
\begin{center}
\begin{small}
\begin{sc}
\begin{tabular}{lcccr}
\toprule
Method & Diffusion Model  & Action Prediction  \\
\midrule
DIAMOND    & $p(s_{t+1}|s_{t:t+T-1}, a_{t:t+T-1})$  & REINFORCE \cite{10.1007/BF00992696} \\
PolyGrad & $p(r_{t:t+T-1}, s_{t+1:t+T-1}|s_t, a_{t:t+T-1})$& stochastic Langevine dynamics \\
PGD    & $p(s_{t:t+T}, a_{t:t+T-1}, r_{t:t+T-1}$) & extract from the sample \\
Diffuser    &  $p(a_{t:t+T-1}, s_{t+1:t+T-1}|s_t)$ & extract from the sample \\
DD     & $p(s_{t+1:t+T-1}|s_t, g_t)$& inverse dynamic models  \\
DWM      & $p(r_{t:t+T-1}, s_{t+1:t+T-1}|s_t, a_t, g_t)$ & MF offline methods \\
D-MPC & $p(s_{t+1:t+T-1}|s_{0:t}, a_{0:t+T-1})$ & diffusion generation  \\
\textbf{DAWM(ours)}      & $p(s_{t+1:t+T-1}, r_{t:t+T-1}|s_t, a_t, g_t)$      & inverse dynamic methods \\

\bottomrule
\end{tabular}
\end{sc}
\end{small}
\end{center}
\vskip -0.1in
\end{table*}
Existing diffusion-based methods for offline RL generally fall into two categories. Value-based approaches such as the Diffusion World Model (DWM) \cite{ding2024diffusionworldmodelfuture} enables value-based RL through a model-based value expansion (MVE) framework \cite{feinberg2018modelbasedvalueestimationefficient}, which estimates target value using action-free trajectories generated by a diffusion model. However, an empirical analysis by \cite{palenicek2022revisitingmodelbasedvalueexpansion} reveals that multi-step value expansion methods, may offer diminishing returns even when the dynamics model is perfectly accurate. This suggests that the limitations arise not only from the model error but also from the structure of long-horizon target value estimation itself. Moreover, because DWM does not explicitly model actions, it is incompatible with one-step TD learning \cite{articletd}, which requires consistency between states, actions, and rewards for each transition. One-step TD learning has been widely adopted in offline RL due to its ability to improve critic regularization and mitigate overfitting \cite{eysenbach2023connectiononestepregularizationcritic}. These insights highlight the importance of generating complete transitions that support stable TD learning. A natural alternative is to directly generate complete $(s, a, r)$ trajectories in an end-to-end fashion, but the prior work \cite{zhou2024diffusionmodelpredictivecontrol, ajay2023conditionalgenerativemodelingneed} has shown that such joint modeling can suffer from significant training instability, limiting its effectiveness in practice.

On the other hand, planning-based methods such as Decision Diffuser (DD) \cite{ajay2023conditionalgenerativemodelingneed}, Diffuser \cite{janner2022planningdiffusionflexiblebehavior}, and D-MPC \cite{zhou2024diffusionmodelpredictivecontrol} generate complete trajectories by recovering actions via an IDM (DD), jointly modeling states and actions through diffusion (Diffuser), or training a separate diffusion-based action planner (D-MPC). These methods bypass TD learning by generating full trajectories and selecting them post hoc using reward signals obtained from environment simulators, instead of learning a value function. However, they require generating entire trajectories during inference and typically rely on iterative sampling and reward-based selection, resulting in a high inference cost. This issue is especially pronounced in D-MPC, which trains a separate diffusion model solely for action generation.

Motivated by the limitations of both paradigms, our goal is to enable one-step TD learning in offline RL using data generated by a diffusion-based model that is also computationally efficient. To achieve this, we adopt a modular approach that generates state-reward sequences via diffusion and infers actions through a separately trained IDM. This modular design yields trajectories with complete transitions $(s, a, r, s')$ for offline RL training, while enabling efficient action inference via a single forward pass, in contrast to diffusion-based planners that require iterative sampling and are computationally expensive. To this end, we introduce \textbf{Diffusion Action World Model (DAWM)}, which bridges efficient generative modeling and TD-based policy learning in offline RL.


To summarize, we pose the following research question:

\textit{Can incorporating action inference into diffusion world models enhance the utility of synthetic data for offline RL, and does it lead to measurable and efficient performance gains over models that rely on action-free or reward-free trajectories?}

To answer this question, we construct synthetic rollouts by combining diffusion-based state-reward generation conditioned on $(s_t, a_t, R_{\text{tg}})$ with IDM-based action inference. We show empirically that the offline RL algorithms TD3BC \cite{fujimoto2021minimalist} and IQL \cite{kostrikov2021offlinereinforcementlearningimplicit} benefit significantly from training on these generated trajectories, achieving consistent performance improvements over prior diffusion-based baselines. These results highlight the importance of action inference in improving the utility of diffusion world models for offline reinforcement learning.

Our main contributions include:
\begin{itemize}
    \item We propose an efficient and effective modular framework that combines diffusion-based world models with IDM to generate full transitions for offline reinforcement learning.
    \item We show that conservative offline RL algorithms can effectively leverage these generated transitions, achieving improved performance over existing diffusion-based baselines.
    \item Our results demonstrate that DAWM-generated trajectories can closely match the performance of agents trained on real offline datasets, highlighting the potential of synthetic trajectories as a practical alternative for offline RL training.

\end{itemize}

\section{Related Work}
\textbf{Offline Reinforcement Learning} enables policy learning from fixed datasets without further environmental interaction. While this setting is appealing for real-world applications where exploration is expensive or unsafe, it introduces a fundamental reliance on the quality of available data. In particular, two aspects of data quality are crucial: coverage, which refers to how well the dataset spans the relevant state-action space, and completeness, which refers to whether each trajectory contains fully labeled transitions with states, actions, and rewards. A key challenge in this setting is the extrapolation error \cite{fujimoto2019offpolicydeepreinforcementlearning}, which occurs when the Q-function is queried on poorly covered or unseen state-action pairs, often leading to overestimation and unstable policy updates \cite{kumar2020conservativeqlearningofflinereinforcement}. To mitigate this, various conservative and behavior-regularized algorithms have been proposed. Methods such as BCQ \cite{fujimoto2019offpolicydeepreinforcementlearning} and TD3BC restrict the learned policy to remain close to the behavior policy, while CQL \cite{kumar2020conservativeqlearningofflinereinforcement} penalizes value estimates on unseen actions. IQL avoids extrapolation entirely by learning value-weighted policies from in-distribution actions only. All of these methods assume access to high-coverage complete datasets. However, in many realistic scenarios, such datasets may be incomplete or difficult to obtain. This motivates our approach: Generate synthetic trajectories that are both fully observed and targeted toward the relevant state-action regions, thereby supporting more stable and effective offline RL training.
\textbf{World Models} \cite{https://doi.org/10.5281/zenodo.1207631} learn representations of environment dynamics, enabling trajectory prediction and planning. Recently, diffusion-based world models have shown strong capabilities in synthesizing realistic sequential data, offering a promising direction for generating trajectories in offline RL. DWM \cite{ding2024diffusionworldmodelfuture}, PGD \cite{jackson2024policyguideddiffusion}, and PolyGRAD \cite{rigter2024worldmodelspolicyguidedtrajectory} generate future states and rewards under different conditions, but PGD and PolyGRAD rely on on-policy rollouts guided by a learned actor, limiting their offline applicability. The DWM paper also proposes a Diffusion-MVE approach, where given an offline state-action pair \((s_t, a_t)\) and return-to-go value $R_{tg}$, the model generates action-free rollouts of states and rewards to compute multi-step returns as TD targets, with bootstrapping applied at the final state. However, the lack of action generation prevents the construction of full transitions \((s, a, r, s')\), making it incompatible with one-step TD learning. Diffuser \cite{janner2022planningdiffusionflexiblebehavior} and DD \cite{ajay2023conditionalgenerativemodelingneed} also generate future trajectories but differ in usage: Diffuser jointly models states and actions via an unconditional model, while DD predicts future states conditioned on \(s_t\) and return-to-go value $g_t$, recovering actions via an IDM. Both rely on post hoc evaluation using environment reward queries, rather than incorporating generated data into RL training. In contrast, our approach generates complete trajectories with states, actions, and rewards, enabling direct integration with one-step TD-based offline RL.


\textbf{IDM} have been used in exploration to learn representations of controllable aspects of the state \cite{pathak2017curiositydrivenexplorationselfsupervisedprediction}. In offline RL, IDMs have been used to recover missing actions from state transitions \cite{torabi2018behavioralcloningobservation}. Behavioral Cloning \cite{torabi2018behavioralcloningobservation} demonstrated the potential of IDMs by learning actions directly from state-only trajectories. However, this framework is limited to imitation learning and does not generalize to offline RL settings. Li et al. \cite{li2023robustvisualimitationlearning} extended IDM to visual control tasks, incorporating it as a regularization objective to improve robustness in imitation learning, but their method is specifically designed for visual inputs, and it is less applicable to state-based offline RL. More recently, SSORL \cite{zheng2023semisupervisedofflinereinforcementlearning} leveraged IDM to infer actions to construct complete trajectories, effectively transforming state-only data into fully labeled data for semi-supervised offline RL. We adopt the SSORL IDM component as it generalizes to action-free data in state-based settings, aligning with our objective of converting action-free diffusion-generated trajectories into complete trajectories for one-step TD learning.

\begin{algorithm}[tb]
   \caption{\textbf{D}iffusion \textbf{A}ction \textbf{W}orld \textbf{M}odel for Offline RL}
   \label{alg:dawm}
\begin{algorithmic}
   \STATE {\bfseries Input:} pretrained diffusion world model $p_{\theta}$, IDM model $f_{\phi}$, offline RL data $D$
   \STATE {\bfseries Hyperparameters:} rollout length $H$, conditioning RTG $g_{eval}$, guidance parameter $\omega$, actor and target networks update frequency $n$, target network update coefficient $\tau$
   
   \STATE \textbf{Initialize} the actor and critic networks $\pi_{\psi}$, $Q_{\Psi}$
   \STATE \textbf{Initialize} the target actor and critic networks $\psi'\leftarrow\psi$, $\Psi'\leftarrow\Psi$
   \FOR{$i=1, 2, ...$ {until convergence}}

   \STATE Sample state-action pair $(s_t, a_t)$ from $D$
   \STATE Sample $\hat{r}_t, \hat{s}_{t+1}, \hat{r}_{t+1},..., \hat{s}_{t+H}, \hat{r}_{t+H},  \hat{s}_{t+H+1}$ $ \sim $

\hspace{1em}$p_{\theta}(\cdot|s_t, a_t, g_{eval})$ with guidance parameter $\omega$
   \STATE Inference actions using $f_{\phi}$ to construct $\mathcal{T}_{\text{complete}}$ $\sim$
   
   \hspace{1em}$\{(s_t, a_t, \hat{r}_t, \hat{s}_{t+1}), ..., (\hat{s}_{t+H}, \hat{a}_{t+H}, \hat{r}_{t+H}, \hat{s}_{t+H+1})\}$
   \FOR{each transition $(s, a, r, s')$ in $\mathcal{T}_{\text{complete}}$}
    
    \STATE Compute the target Q value:
    \STATE \hspace{1em} $y = r + \gamma Q_{\Psi}(s', \pi_{\psi}(s'))$
    
    \STATE Update Critic Networks:
    \STATE \hspace{1em} $\Psi \leftarrow \Psi - \eta_\Psi \nabla_{\Psi} \left( Q_{\Psi}(s, a) - y \right)^2$
    
    \IF{$i \bmod n = 0$}
        \STATE Update Actor Network:
        \STATE \hspace{1em} $\psi \leftarrow \psi - \eta_\psi \nabla_\psi \left( Q_{\Psi}(s, \pi_\psi(s)) \right)$
        
        \STATE Update Target Actor-Critic Networks:
        \STATE \hspace{1em} $\Psi' \leftarrow \tau \Psi + (1 - \tau) \Psi'$
        \STATE \hspace{1em} $\psi' \leftarrow \tau \psi + (1 - \tau) \psi'$
    \ENDIF
\ENDFOR




    


   \ENDFOR

\end{algorithmic}
\end{algorithm}

\section{Diffusion Action World Model}
In this paper, we propose \textbf{Diffusion Action World Model (DAWM)}, a structured framework that consists of two key phases: (1) data synthesis, which constructs complete training trajectories from a conditional diffusion world model $p_{\theta}$ combined with an IDM $f_{\phi}$, and (2) offline policy training, which learns a policy from the synthesized data.

Algorithm~\ref{alg:dawm} outlines the overall framework of our method. In the first phase, we train the Diffusion World Model $p_{\theta}$ to model long horizon transitions and rewards. This model is used to generate synthetic trajectories comprising future states and rewards, but without corresponding actions. To complete these trajectories, we apply a separately trained IDM $f_{\phi}$ to reconstruct the missing actions, yielding full $(s, a, r, s')$ transitions. In the second phase, the generated trajectories are used to train an offline RL agent using standard value-based algorithms. In our experiments, we adopt TD3BC and IQL. Details of these two algorithms can be found in Appendix \ref{apdxalgo}.

\subsection{Conditional Diffusion Trajectory Generation}

As the first stage of DAWM, we adopt the conditional diffusion generation process introduced in DWM, where a diffusion model is trained to generate future sequences of states and rewards, conditioned on the current state $s_t$, action $a_t$, and a target return-to-go value $R_{\text{tg}}$. The model defines the following conditional distribution over sub-trajectories of horizon $H$:

\begin{equation}
p_{\theta}(\hat{s}_{t+1:t+H+1}, \hat{r}_{t:t+H} \mid s_t, a_t, R_{\text{tg}})
\end{equation}

$R_{\text{tg}}$ is calculated as a discounted cumulative reward, $R_{\text{tg}} = \sum_{k=0}^{H} \gamma^k \hat{r}_{t+k}$, using a fixed discount factor $\gamma$. To facilitate stable training across tasks with varying reward scales, we normalize $R_{\text{tg}}$ by dividing it by a constant specific to each environment, chosen based on empirical return statistics from the dataset. More details are provided in Appendix \ref{apdxa}.

Following previous work \cite{ding2024diffusionworldmodelfuture}, the conditional diffusion model is implemented as a U-Net backbone \cite{ronneberger2015unetconvolutionalnetworksbiomedical}, and the model is trained using a scoring denoising objective \cite{6795935, JMLR:v6:hyvarinen05a} with classifier-free guidance \cite{ho2022classifierfreediffusionguidance}.
During pre-training, we minimize the expected error between predicted and true noise under randomly masked conditioning:

\begin{equation}
\label{scorematching}
    \mathbb{E}_{(x^{(0)}, y), k, \epsilon, b}
    \left\|
    \epsilon_{\theta}\left(
    x^{(k)}(x^{(0)}, \epsilon),\,
    k,\,
    (1 - b) \cdot y + b \cdot \varnothing
    \right)
    - \epsilon
    \right\|^2_2
\end{equation}

Here, $x^{(0)}$ denotes a trajectory segment consisting of future interleaved states-rewards pairs, while the conditioning variable $y$ = $(s_t, a_t, R_{tg})$ includes the initial state-action pair and the return-to-go signal. $\epsilon \sim \mathcal{N}(0, I)$ is the added Gaussian noise, $\epsilon_{\theta}(\cdot)$ is the noise prediction network, $k$ stands for the diffusion timestep, and $b \in \{0, 1\}$ is a Bernoulli variable controlling whether the conditioning is applied. In our setting, states are low-dimensional proprioceptive vectors that capture physical attributes such as joint positions and velocities, without any image-based input modalities, and rewards are scalar values. This setup produces future trajectories that are consistent with the desired return but do not contain any action information, which simplifies the generative modeling task by avoiding the complexities associated with jointly modeling actions within the diffusion process. We reconstruct actions using a lightweight IDM for efficient action sequence generation instead of using a costly diffusion-based action planner.


\subsection{Action Completion via IDM}
After generating action-free trajectories, we reconstruct actions using the IDM \cite{zheng2023semisupervisedofflinereinforcementlearning}, thus generating full trajectories that can be directly used by conventional offline RL methods. We follow the stochastic multi-transition IDM proposed by \cite{zheng2023semisupervisedofflinereinforcementlearning}, which models $f_{\phi}(a_t \mid s_{t-m:t-1}, s_t, s_{t+1})$ as a diagonal Gaussian distribution, where $m$ denotes the length of the historical state context used for action inference. The model is trained by maximizing the likelihood of observed actions within the subsequence of trajectories from the real dataset that contain action annotations:
\begin{equation}
    \max_{\phi} \sum_{(s_{t-m:t-1}, s_t, s_{t+1}) \in \mathcal{D}_{\text{DWM}}} \log f_{\phi}(a_t \mid s_{t-m:t-1}, s_t, s_{t+1})
\end{equation}
 At inference time, $f_{\phi}$ is applied to the generated trajectories from the diffusion world model to reconstruct actions. This yields complete $(s, a, r, s')$ tuples.

\subsection{Offline Policy Training Using Reconstructed Trajectories}
As the second stage of DAWM, after constructing complete $(s, a, r, s')$ tuples, the synthesized trajectories are used to train offline RL agents. We consider two representative methods: TD3BC, which augments the TD3 policy objective with a behavior cloning regularizer to keep the policy close to the data, and IQL, which avoids explicit behavior cloning by performing advantage-weighted supervised regression toward high-value actions. In both cases, the Q-functions are trained using Bellman targets \cite{712192} computed from the reconstructed trajectories, where the IDM inferred actions enable one-step temporal difference updates over the model-generated data. The policies are then optimized using their respective update rules. This setup allows us to evaluate how well diffusion-generated data supports policy learning under different offline RL algorithms.
\begin{table*}[t]
\caption{DAWM outperforms DWM and DD on normalized return across multiple locomotion benchmarks.}
\label{tab:dwm}
\vskip 0.15in
\begin{center}
\begin{small}
\begin{sc}
\begin{tabular}{lccccr}
\toprule
Env. & DD & DWM-TD3BC & DAWM-TD3BC & DWM-IQL& DAWM-IQL\\
\midrule
hopper-m    & 0.49$\pm$ 0.07 & 0.65$\pm$ 0.10&\textbf{0.69$\pm$ 0.10}& 0.54$\pm$ 0.11& \textbf{0.64$\pm$0.13}\\
hopper-mr & 0.66$\pm$ 0.15 & 0.53$\pm$ 0.09&\textbf{0.79$\pm$0.01}&0.61$\pm$ 0.13&\textbf{0.64$\pm$0.07}\\
hopper-me & 1.06$\pm$ 0.11 & 1.03$\pm$ 0.14&\textbf{1.08$\pm$0.07}&0.82$\pm$ 0.23&\textbf{0.88$\pm$0.20}\\
\midrule
walker2d-m & 0.67$\pm$ 0.16 & 0.70$\pm$ 0.15&\textbf{0.79$\pm$ 0.12}&0.76$\pm$ 0.50& \textbf{0.85$\pm$0.05}\\
walker2d-mr & 0.44$\pm$ 0.26 & 0.46$\pm$ 0.19&\textbf{0.61$\pm$0.13}&0.35$\pm$ 0.14&\textbf{0.54$\pm$0.16}\\
walker2d-me & 0.99$\pm$ 0.15 & \textbf{1.10$\pm$ 0.00}&\textbf{1.10$\pm$ 0.00}&1.04$\pm$ 0.10&\textbf{1.10$\pm$0.00}\\
\midrule
halfcheetah-m   & \textbf{0.49$\pm$ 0.01} & 0.46$\pm$ 0.01& 0.47$\pm$ 0.01&0.44$\pm$ 0.01&0.44$\pm$0.01\\
halfcheetah-mr& 0.38$\pm$ 0.06 & 0.43$\pm$ 0.01&\textbf{0.44$\pm$0.01}&0.40$\pm$ 0.01&\textbf{0.41$\pm$0.01}\\
halfcheetah-me& \textbf{0.91$\pm$ 0.01} &0.75$\pm$ 0.16&0.71$\pm$0.23 &0.71$\pm$ 0.14&0.71$\pm$0.22\\
\midrule
averge &0.677$\pm$0.109 & 0.679$\pm$0.098&\textbf{0.742$\pm$0.076}&0.630$\pm$0.117&\textbf{0.690 $\pm$0.094}\\
\bottomrule
\end{tabular}
\end{sc}
\end{small}
\end{center}
\vskip -0.1in
\end{table*}

\begin{table}[t]
\caption{Inference time (s) comparison between DD, DWM-TD3BC, and DAWM-TD3BC across multiple locomotion benchmarks. }
\label{tab:dawmtime}
\vskip 0.15in
\begin{center}
\begin{small}
\begin{sc}
\resizebox{0.95\linewidth}{!}{
\begin{tabular}{lcccr}
\toprule
Env.  & DD & DWM-TD3BC & DWAM-TD3BC \\

\midrule
hopper-m    & 4.11s$\pm$ 4.64s& 1.18s$\pm$ 0.51s&  1.19s$\pm$ 0.48s\\
hopper-mr & 6.21s$\pm$ 4.21s& 0.64s$\pm$ 0.52s&0.64s$\pm$ 0.53s\\
hopper-me & 8.82s$\pm$ 2.96s& 2.62s$\pm$ 1.22s& 2.63s$\pm$ 1.18s\\
\midrule
walker2d-m & 8.09s$\pm$ 1.24s& 2.00s$\pm$ 0.63s& 2.02s$\pm$ 0.61s\\
walker2d-mr & 5.94s$\pm$ 4.32s& 0.83s$\pm$ 0.45s&0.84s$\pm$ 0.51s\\
walker2d-me & 9.26s$\pm$ 1.30s& 3.60s$\pm$ 3.90s&3.61s$\pm$ 3.91s\\
\midrule
halfcheetah-m   & 8.18s$\pm$ 3.77s & 1.81s$\pm$ 0.54s& 1.83s$\pm$ 0.52s \\
halfcheetah-mr& 7.46s$\pm$ 9.72s& 0.60s$\pm$ 0.17s&0.61s$\pm$ 0.19s\\
halfcheetah-me& 9.53s$\pm$ 2.50s& 3.77s$\pm$ 2.43s&3.78s$\pm$ 2.51s\\
\midrule
Average &7.531s $\pm$ 3.651s &1.894s$\pm$1.379s&1.904s$\pm$1.301s\\
\bottomrule

\end{tabular}}
\end{sc}
\end{small}
\end{center}
\vskip -0.1in
\end{table}

\section{Experiments}
In our experimental evaluation, we examine the following key research questions:

\begin{itemize}
    \item To what extent can action completion via IDM enhance the effectiveness of diffusion-generated trajectories for offline RL?
    \item Can policies trained on DAWM-generated data achieve close performance compared to those trained directly on real offline trajectories?
    \item Does the amount of synthetic data and the generation horizon influence the effectiveness of DAWM in reinforcing one-step TD learning?
\end{itemize}

First, to assess whether diffusion-generated trajectories with IDM-reconstructed actions can serve as more effective supervision for offline RL, we select DWM as a baseline. We adpot TD3BC and IQL on top of differently generated trajectories, namely on top of DWM and DAWM to compare their performance. Note that DD reconstructs actions using an IDM from diffusion-based generated states-only trajectory to do the planning for different tasks directly, making it a complementary baseline. 
Next, to assess how closely policies trained on DAWM-generated data can match those trained on real offline trajectories, we compare DAWM to two baselines: standard TD3BC and IQL trained on standard offline data, and the SSORL agents. SSORL framework leverages real action-free trajectories and applies an IDM to reconstruct missing actions, serving as a baseline for action completion on real data. Finally, we introduce DAWM-T, a variant that samples $1/T$ of DAWM’s conditioning $(s_t, a_t)$ pairs, but with the same length of diffusion rollouts, reducing both the number of transitions and generated trajectories by a factor of $T$, allowing us to isolate the impact of data volume while retaining the same action completion strategy.

\textbf{Benchmark and Implementation Details:} We conduct experiments to evaluate our DAWM framework on 9 standard D4RL \cite{fu2020d4rl} locomotion tasks, and report the obtained normalized return (0-1) as evaluation metrics, where 1 indicates expert performance. The normalized return is calculated as: $$\text{Normalized Return} = \frac{R_{\text{agent}} - R_{\text{random}}}{R_{\text{expert}} - R_{\text{random}}}$$

where $R_{\text{agent}}$ is the average return achieved by the agent, and $R_{\text{random}}$, $R_{\text{expert}}$ denote returns of the random policy and expert policy defined for the environment in the D4RL benchmark. We set the trajectory length generated by DAWM to \( H = 7 \), where each trajectory begins with a real state-action pair \((s_t, a_t)\) from the offline dataset and is followed by 7 transitions \((s, a, r)\) generated by the model.
We set $k=5$ as diffusion steps for the diffusion world model training. For DAWM-T, we select $T = 8$, which indicates DAWM-T generates 8 times less data compared to DAWM. More details can be found in Appendix \ref{apdxa}.

\begin{table*}[t]
\caption{DAWM-trained agents on fully synthetic data achieve comparable or better normalized return than  SSORL-trained agents using action-completed datasets, and outperform TD3BC/IQL trained on real data.}
\label{tab:ssorl}
\vskip 0.15in
\begin{center}

\begin{small}
\begin{sc}

\begin{tabular}{lcccccr}
\toprule
Env. & TD3BC & SSORL-TD3BC & DAWM-TD3BC & IQL &SSORL-IQL & DAWM-IQL   \\

\midrule
hopper-m    & 0.58$\pm$ 0.11& 0.64$\pm$ 0.11&\textbf{0.69$\pm$ 0.10} & 0.61$\pm$ 0.06&\textbf{0.69$\pm$0.22}&0.64$\pm$0.13\\
hopper-mr & 0.53$\pm$ 0.19& 0.62 $\pm$0.25&\textbf{0.79$\pm$ 0.01}& \textbf{0.87$\pm$ 0.16}&0.77$\pm$0.31&0.64$\pm$0.07\\
hopper-me & 0.9$\pm$ 0.28& 0.96 $\pm$ 0.25&\textbf{1.08$\pm$ 0.07}&0.77$\pm$ 0.35&\textbf{1.10$\pm$0.01} &0.88 $\pm$0.20\\ 
\midrule
walker2d-m & 0.77$\pm$ 0.09& \textbf{0.81$\pm$ 0.05}&0.79$\pm$ 0.12&0.79$\pm$ 0.05& \textbf{0.85$\pm$0.01}&0.85$\pm$0.05\\
walker2d-mr & \textbf{0.75$\pm$ 0.19}& 0.73 $\pm$ 0.27&0.61$\pm$ 0.13&0.68$\pm$ 0.06&\textbf{0.92$\pm$0.04} &0.54$\pm$0.16\\
walker2d-me & 1.08$\pm$ 0.01& \textbf{1.11$\pm$0.03}&1.10$\pm$ 0.00&1.08$\pm$ 0.02&\textbf{1.11$\pm$0.35} &1.10$\pm$0.00\\
\midrule
halfcheetah-m   & 0.47$\pm$ 0.01 & \textbf{0.48$\pm$ 0.01}& 0.47$\pm$ 0.01 &\textbf{0.48$\pm$ 0.00}&\textbf{0.48$\pm$0.01}&0.44$\pm$0.01\\
halfcheetah-mr& 0.43$\pm$ 0.01& \textbf{0.44 $\pm$0.01}& \textbf{0.44 $\pm$0.01}&0.43$\pm$ 0.02&\textbf{0.45$\pm$0.01}&0.41$\pm$0.01\\
halfcheetah-me& 0.73$\pm$ 0.16&\textbf{0.90$\pm$ 0.03}&0.71$\pm$ 0.23&0.88$\pm$ 0.03&\textbf{0.94$\pm$0.01}&0.71$\pm$0.22\\
\midrule
average &0.693$\pm$0.116&\textbf{0.743$\pm$0.112}&0.742$\pm$0.076&0.732$\pm$0.083&\textbf{0.812 $\pm$0.108}&0.690 $\pm$0.094\\
\bottomrule

\end{tabular}

\end{sc}
\end{small}
\end{center}
\vskip -0.1in
\end{table*}

\begin{table}[t]
\caption{Normalized return comparison between DAWM-T and DAWM across 9 locomotion benchmarks. }
\label{tab:dawmt}
\vskip 0.15in
\begin{center}
\begin{small}
\begin{sc}

\begin{tabular}{lcccr}
\toprule
Env.  & DAWM-T & DAWM \\
\midrule
hopper-m   & 0.73$\pm$ 0.16&0.69$\pm$ 0.10&  \\
walker2d-m & 0.77$\pm$ 0.06&0.79$\pm$ 0.12& \\
halfcheetah-m    & 0.44$\pm$ 0.01& 0.47$\pm$ 0.01& \\
\bottomrule
\end{tabular}
\end{sc}
\end{small}
\end{center}
\vskip -0.1in
\end{table}

\begin{table*}[t]
\caption{Normalized return of DAWM and DWM with varying generation horizons ($H = 1, 3, 7$)}
\label{tab:horizon}
\vskip 0.15in
\begin{center}
\begin{small}
\begin{sc}

\begin{tabular}{lccccccr}
\toprule
Env. & DAWM\_1 &DWM\_1 & DAWM\_3 &DWM\_3& DAWM\_7&DWM\_7 \\
\midrule
hopper-m    &\textbf{0.77$\pm$ 0.10}&0.68$\pm$ 0.12&  \textbf{0.74$\pm$ 0.09}&0.63$\pm$ 0.11&\textbf{0.69$\pm$ 0.10}&0.65$\pm$ 0.10 \\
walker2d-m &\textbf{0.76$\pm$ 0.08}&0.56$\pm$ 0.13& 0.73$\pm$ 0.06&\textbf{0.74$\pm$ 0.13}&\textbf{0.79$\pm$ 0.12}&0.70$\pm$ 0.15 \\
halfcheetah-m   &\textbf{0.45$\pm$ 0.09}&0.35$\pm$ 0.03& \textbf{0.44$\pm$ 0.10}&0.39$\pm$ 0.01& \textbf{0.47$\pm$ 0.01}& 0.46$\pm$ 0.01\\
\bottomrule
\end{tabular}
\end{sc}
\end{small}
\end{center}
\vskip -0.1in
\end{table*}

\subsection{Action Completion as Enhanced Supervision for One-Step TD Learning}
To evaluate the role of action completion in offline reinforcement learning, we compare DAWM with DWM. We
report results using DAWM and DWM with $H = 7$. By incorporating an IDM to infer missing actions, DAWM generates complete transitions that are compatible with one-step temporal-difference (TD) learning. In Table \ref{tab:dwm}, experimental results across all nine D4RL locomotion benchmarks show that DAWM consistently outperforms DWM, achieving an average improvement of 9.3\% with the TD3BC agent and 9.5\% with the IQL agent. These results highlight the benefits of complete state-action-reward transitions in supporting more effective policy learning.

The advantage of action supervision is particularly evident in Walker2d, a locomotion task characterized by more complex dynamics and greater sensitivity to control compared to HalfCheetah \cite{szulc2020frameworkreinforcementlearningautocorrelated, furuta2021policyinformationcapacityinformationtheoretic}, where DAWM demonstrates substantial gains over DWM. In contrast, the improvement observed in HalfCheetah is relatively marginal. These findings suggest that action-complete trajectories may be particularly beneficial in environments where control is more challenging, although further investigation is needed in more diverse or complex settings. 

We also compare DAWM with DD, a diffusion-based trajectory planner. Although the two approaches differ in their training formulations, we evaluate both using the same performance metrics. In terms of normalized return, DAWM achieves superior results under both TD3BC and IQL agents, demonstrating that DAWM provides a more effective and scalable alternative to diffusion-based planning methods such as DD. Furthermore, as shown in Table~\ref{tab:dawmtime}, DAWM achieves approximately 4.0× faster inference, underscoring its practicality and efficiency for large-scale generative offline RL applications.

Overall, these findings suggest that DAWM effectively and efficiently leverages action information to provide stronger learning signals in environments with high transition variability, and its advantages appear more pronounced in tasks exhibiting greater control complexity, as observed in D4RL benchmarks.

\subsection{Comparison to Agent Trained on Real Offline Dataset}
To assess the effectiveness of trajectories generated by DAWM compared to real offline datasets, we evaluate the performance of TD3BC and IQL agents trained on DAWM-generated data against those trained on real datasets using both standard and SSORL-augmented protocols. With $H = 7$, the DAWM-generated dataset contains approximately eight times more transitions than the original datasets. As shown in Table~\ref{tab:ssorl}, DAWM achieves comparable average returns across both RL agents, outperforming its standard counterpart TD3BC by 7.1\% and showing competitive results with IQL. Although it underperforms SSORL agents, this may be partially explained by the fact that SSORL applies a data filter that selects the top-K longest trajectories from the offline dataset, while DAWM generates fixed-length trajectories without trajectory length selection. This discrepancy may affect the training distribution, as longer trajectories are more likely to contain higher-return transitions, which are preferentially selected by SSORL but not by DAWM. These findings suggest that DAWM-generated data offer a strong training signal and can serve as a viable alternative to real offline datasets. Moreover, while the larger dataset size may contribute to the observed gains, we further isolate this factor by conducting controlled experiments on a subset of three environments, where TD3BC is trained under varying dataset sizes.



\subsection{Data Volume Analysis: DAWM-T vs. DAWM}

We introduce DAWM-T, a variant that maintains the same training procedure as DAWM but reduces the number of sampled conditioning pairs during inference to $1/T$ of DAWM's sampling volume. Consequently, both the number of conditioning pairs and the number of generated trajectories are scaled down by a factor of $T$. This setting allows us to assess to what extent the performance gains of DAWM can be attributed to the amount of generated data. 

Table~\ref{tab:dawmt} presents the performance of TD3BC agents trained using DAWM-T and DAWM generated data, allowing for a direct comparison of data volume effects under identical training precedure, we set $T = 8$. DAWM achieves higher returns in Walker2d and HalfCheetah, while DAWM-T performs slightly better in Hopper. Overall, the performance differences are relatively modest and vary across environments. These results provide evidence that action completion may be a more influential factor than data volume in the settings we evaluated.

\subsection{Generation Horizon Analysis: DAWM with Different Generation Horizon}
Table~\ref{tab:horizon} compares the performance of DAWM-TD3BC and DWM-TD3BC across three generation horizons ($H = 1$, $H = 3$, $H = 7$) on the locomotion benchmark. We observe two consistent patterns. First, DAWM outperforms DWM under all horizon settings, highlighting the importance of action completion for generating effective training data. Second, the performance of both DAWM and DWM remains relatively stable across different rollout lengths, with only minor fluctuations. This suggests that the DWM produces coherent predictions even with short horizons, and that our one-step TD learning framework is robust to the choice of generation length. Together, these results demonstrate that DAWM achieves stronger performance than prior DWM approaches while retaining flexibility in data generation.

\section{Conclusion}
In this paper, we propose DAWM, a modular framework that We introduced DAWM, a modular world model that combines diffusion-based state-reward generation with inverse dynamics-based action completion to synthesize fully labeled trajectories for offline reinforcement learning. By enabling one-step TD learning on synthetic transitions, DAWM offers a simple yet effective alternative to prior diffusion-based approaches, delivering both stronger policy performance and lower inference cost. Notably, policies trained on DAWM-generated data not only outperform those trained on prior generative baselines, and closes the performance gap with agents trained directly on the original offline datasets, demonstrating the potential of synthetic trajectories as a viable alternative.

Looking forward, we plan to extend DAWM to more complex domains such as visual control and high-dimensional robotics, as well as investigate its robustness under imperfect dynamics and action inference. Exploring hybrid pipelines that integrate DAWM-generated data with online fine-tuning is another promising direction.

\section*{Acknowledgements}

The authors gratefully acknowledge the scientific support and HPC resources provided by the Erlangen National High Performance Computing Center (NHR@FAU) of the Friedrich-Alexander-Universität Erlangen-Nürnberg (FAU) under the NHR project b241dd10. NHR funding is provided by federal and Bavarian state authorities. NHR@FAU hardware is partially funded by the German Research Foundation (DFG) – 440719683.

We also thank the Munich Center for Machine Learning (MCML) for supporting this research through access to GPU machines and for partially funding conference participation.

\bibliography{example_paper}
\bibliographystyle{icml2025}

\newpage
\appendix
\onecolumn
\section{Appendix}
\subsection{Hyperparameters}
\label{apdxa}
We summarize the architectures and all hyperparameters used in our experiments in this section. For all experiments, we use our own PyTorch implementation of the Diffusion World Model (DWM). For the Inverse Dynamics Model (IDM), TD3+BC, and IQL, our implementations are based on the following public codebases:

\begin{itemize}
    \item \textbf{SSORL}: \url{https://github.com/facebookresearch/ssorl}
    \item \textbf{TD3+BC}: \url{https://github.com/sfujim/TD3_BC}
    \item \textbf{IQL}: \url{https://github.com/gwthomas/IQL-PyTorch}
\end{itemize}

Table~\ref{hy-dwm} summarizes the key hyperparameters used during training. The diffusion world model was trained for 2 million steps using the Adam optimizer. For inference, we use $N = 3$ as reduced diffusion steps. The probability of null conditioning is set to 0.25. The model follows a U-Net backbone with 6 repeated residual blocks, each containing 2 temporal convolution layers with $Mish$ activation.

For conditional inputs, we used separate projection networks for each conditioning variable: a 2-layer MLP for timestep $K$, and two 3-layer MLPs for return-to-go $R_{\text{tg}}$ and action $a_t$, respectively. The $R_{tg}$ value is normalized by a task-specific value, which is 400 for Hopper, 550 for Walker2d, and 1200 for Halfcheetah tasks.

The inverse dynamics model (IDM) is implemented as two independent 2-layer MLPs with $1024$ neurons per layer, used to predict the mean and covariance of the inferred action distribution. 

\begin{table}[h]
\caption{Hyperparameters of DAWM}
\label{hy-dwm}
\vskip 0.15in
\begin{center}
\begin{small}
\begin{sc}
\begin{tabular}{lcr}
\toprule
Hyperparameter & Value \\
\midrule
DMW training seed   & 100 \\
diffusion steps $K$ & 5\\
discount factor $\gamma$   & 0.99 \\
lr  & 1e-4 \\
batch size & 64 \\
Adam decay & 0.995\\
Sampling Temperature $\alpha$  & 0.5 \\

IDM context length & 1\\
IDM LR & 1e-4\\
\bottomrule
\end{tabular}
\end{sc}
\end{small}
\end{center}
\vskip -0.1in
\end{table}

\newpage

\subsection{DAWM Variantions}
\label{apdxalgo}
Our DAWM-TD3BC and DAWM-IQL are detailed in Algorithm \ref{alg:dawmtd3bc} and Algorithm \ref{alg:dawmtiql}. 
\begin{algorithm}[H]
   \caption{\textbf{D}iffusion \textbf{A}ction \textbf{W}orld \textbf{M}odel for TD3BC}
   \label{alg:dawmtd3bc}
\begin{algorithmic}

\STATE {\bfseries Input:} pretrained diffusion world model $p_{\theta}$, IDM model $f_{\phi}$, offline RL data $D$
\STATE {\bfseries Hyperparameters:} target return $g_{eval}$, guidance parameter $\omega$, TD loss coefficient $\lambda$, actor and target networks update frequency $n$, target network update coefficient $\tau$
\STATE \textbf{Initialize} the actor and critic networks $\pi_{\psi}$, $Q_{\Psi_{1}}$, $Q_{\Psi_{2}}$
\STATE \textbf{Initialize} the target actor and critic networks $\psi'\leftarrow\psi$, $\Psi_{1}'\leftarrow\Psi_1$, $\Psi_{2}'\leftarrow\Psi_2$
\FOR{$i = 1, 2, \dots$ \textbf{until convergence}}
    \STATE Sample state-action pair $(s_t, a_t)$ from $D$
    \STATE Sample $\hat{r}_t, \hat{s}_{t+1}, \hat{r}_{t+1}, \dots, \hat{s}_{t+H+1} \sim p_{\theta}(\cdot \mid s_t, a_t, g_{eval})$ with guidance $\omega$
    \STATE Infer actions $\hat{a}_{t+1}, \dots, \hat{a}_{t+H} = f_{\phi}(\hat{s}_{t+1}, \dots, \hat{s}_{t+H})$
    \STATE Construct $\mathcal{T}_{\text{labelled}} = \left\{ (\hat{s}_{t+h}, \hat{a}_{t+h}, \hat{r}_{t+h}, \hat{s}_{t+h+1}) \right\}_{h=0}^{H}$
    
    \FOR{each $(s, a, r, s') \in \mathcal{T}_{\text{labelled}}$}
        \STATE Compute the target Q value: \\
        \hspace{1em} $y = r + \gamma \min_{j \in \{1, 2\}} Q_{\Psi_j'}(s', \pi_{\psi'}(s'))$
        
        \STATE Update Critic Networks: \\
        \quad$\Psi_1 \leftarrow \Psi_1 - \eta_\Psi \nabla_{\Psi_1} \left( Q_{\Psi_1}(s, a) - y \right)^2$ \\
        \quad$\Psi_2 \leftarrow \Psi_2 - \eta_\Psi \nabla_{\Psi_2} \left( Q_{\Psi_2}(s, a) - y \right)^2$

        \IF{$i \bmod n = 0$}
            \STATE Update Actor Network: \\
            \hspace{1em} $\psi \leftarrow \psi - \eta_\psi \nabla_\psi \left( \lambda Q_{\Psi_1}(s, \pi_\psi(s)) + \lambda \|a - \pi_\psi(s)\|^2 \right)$

            \STATE Update Target Actor-Critic Networks: \\
            \hspace{1em} $\Psi_j' \leftarrow \tau \Psi_j + (1 - \tau) \Psi_j', \quad j \in \{1, 2\}$ \\
            \hspace{1em} $\psi' \leftarrow \tau \psi + (1 - \tau) \psi'$
        \ENDIF
    \ENDFOR
\ENDFOR

\end{algorithmic}
\end{algorithm}

\begin{algorithm}[t]
   \caption{\textbf{D}iffusion \textbf{A}ction \textbf{W}orld \textbf{M}odel for IQL}
   \label{alg:dawmtiql}

\begin{algorithmic}
\STATE {\bfseries Input:} pretrained diffusion world model $p_{\theta}$, IDM model $f_{\phi}$, offline RL data $D$
\STATE {\bfseries Hyperparameters:} rollout length $H$, conditioning RTG $g_{eval}$, guidance parameter $\omega$, actor and target networks update frequency $n$, target network update coefficient $\tau$
\STATE \textbf{Initialize} actor $\pi_\psi$, critics $Q_{\Psi_1}, Q_{\Psi_2}$, value network $V_\xi$
\STATE \textbf{Initialize} target critics $\Psi_1' \leftarrow \Psi_1$, $\Psi_2' \leftarrow \Psi_2$
\FOR{$i = 1, 2, \dots$ \textbf{until convergence}}
    \STATE Sample state-action pair $(s_t, a_t)$ from offline dataset $D$
    \STATE Generate rollout: \\
    \hspace{1em} Sample $\hat{r}_t, \hat{s}_{t+1}, \dots, \hat{s}_{t+H+1} \sim p_\theta(\cdot \mid s_t, a_t, g_{eval})$ with guidance $\omega$ \\
    \hspace{1em} Infer actions $\hat{a}_{t+1}, \dots, \hat{a}_{t+H} = f_\phi(\hat{s}_{t+1}, \dots, \hat{s}_{t+H})$ \\
    \hspace{1em} Construct labeled one-step transitions $\mathcal{T}_{\text{labelled}} = \{(\hat{s}_{t+h}, \hat{a}_{t+h}, \hat{r}_{t+h}, \hat{s}_{t+h+1})\}_{h=0}^{H}$
    
    \FOR{each $(s, a, r, s') \in \mathcal{T}_{\text{labelled}}$}
        \STATE Compute TD target: \\
        \hspace{1em} $y = r + \gamma \cdot V_\xi(s')$
        
        \STATE Update value network: \\
        \hspace{1em} $\xi \leftarrow \xi - \eta \nabla_\xi \left( \min_{i \in \{1,2\}} Q_{\Psi_i'}(s, a) - V_\xi(s) \right)^2$

        \STATE Update critic networks: \\
        \hspace{1em} $\Psi_1 \leftarrow \Psi_1 - \eta_\Psi \nabla_{\Psi_1} \left( Q_{\Psi_1}(s, a) - y \right)^2$ \\
        \hspace{1em} $\Psi_2 \leftarrow \Psi_2 - \eta_\Psi \nabla_{\Psi_2} \left( Q_{\Psi_2}(s, a) - y \right)^2$
        
        \IF{$i \bmod n = 0$}
            \STATE Update actor network:\\
            \hspace{1em}
            $\psi \leftarrow \psi - \eta_\psi \nabla_\psi \left[ \exp\left( \beta \cdot \left( \min_{i} Q_{\Psi_i}(s, \pi_\psi(s)) - V_\xi(s) \right) \right) \cdot \log \pi_\psi(s) \right]$

            \STATE Update target networks: \\
            \hspace{1em} $\Psi_j' \leftarrow \tau \Psi_j + (1 - \tau) \Psi_j', \quad j \in \{1, 2\}$ \\
            \hspace{1em} $\psi' \leftarrow \tau \psi + (1 - \tau) \psi'$
        \ENDIF
    \ENDFOR
\ENDFOR
\end{algorithmic}

\end{algorithm}



\end{document}